\pdfoutput=1
\documentclass[11pt,a4paper]{article}
\usepackage{times,latexsym}
\usepackage{url}
\usepackage[T1]{fontenc}

\usepackage{graphicx}
\usepackage{booktabs}
\usepackage{rotating}

\usepackage[acceptedWithA]{tacl2021v1}
\title{Thai Universal Dependency Treebank}

\author{
  \textbf{Panyut Sriwirote}$^1$
  \textbf{Wei Qi Leong}$^2$
  \textbf{Charin Polpanumas}$^3$
  \\
  \textbf{Santhawat Thanyawong}$^4$
  \textbf{William Chandra Tjhi}$^2$
  \\
  \textbf{Wirote Aroonmanakun}$^1$
  \textbf{Attapol T. Rutherford}$^1$
  \\
  $^1$Department of Linguistics, Chulalongkorn University, Thailand
  \\
  $^2$AI Singapore, Singapore
  $^3$Amazon, Japan
  \\
  $^4${P}rince {o}f {S}ongkhla {U}niversity, Thailand
  \\
  \texttt{\{panyutsriwirote, cebril\}@gmail.com,}
  \texttt{\{weiqi, wtjhi\}@aisingapore.org,}
  \\
  \texttt{santhawat.t@psu.ac.th,}
  \texttt{\{awirote, attapol.t\}@chula.ac.th}
}

\begin{document}

\maketitle

\begin{abstract}
Automatic dependency parsing of Thai sentences has been underexplored, as evidenced by the lack of large Thai dependency treebanks with complete dependency structures and the lack of a published systematic evaluation of state-of-the-art models, especially transformer-based parsers. In this work, we address these problems by introducing Thai Universal Dependency Treebank (TUD), a new largest Thai treebank consisting of 3,627 trees annotated in accordance with the Universal Dependencies (UD) framework. We then benchmark dependency parsing models that incorporate pretrained transformers as encoders and train them on Thai-PUD and our TUD. The evaluation results show that most of our models can outperform other models reported in previous papers and provide insight into the optimal choices of components to include in Thai dependency parsers. The new treebank and every model's full prediction generated in our experiment are made available on a GitHub repository for further study.\iftaclpubformat\footnote{\url{https://github.com/nlp-chula/TUD}}\else\footnote{[redacted for anonymity]}\fi
\end{abstract}

\section{Introduction}

Research on automatic dependency parsing of Thai sentences has long been underdeveloped. According to a survey by \citet{arreerard2022survey}, there exist only two publicly available dependency treebanks for the Thai language, Thai-PUD,\footnote{\url{https://universaldependencies.org/treebanks/th_pud/index.html}} and Blackboard Treebank,\footnote{\url{https://aiforthai.in.th/corpus.php} (registration required)} each with its own shortcomings. Thai-PUD consists of only 1,000 sentences, which dwarfs in comparison to treebanks of high-resource languages, such as the Universal Dependencies English Web Treebank \citep{silveira2014big_english_treebank}, which consists of 16,622 sentences. On the other hand, Blackboard Treebank consists of 130,561 clauses, making the treebank impressively large, but it suffers from the fact that subordinate and relative clauses are detached and analyzed separately from main clauses, making each parsing unit relatively short and simple. Moreover, in regard to dependency structure, the dependency arcs in Blackboard Treebank are annotated without types. That is, each arc has no associated type that specifies the nature of the relation between the head and the dependent. These factors mean that, if Blackboard Treebank were used as training data, the trained parsers would have limited capability in analyzing full, complex sentences. In fact, they would not be able to distinguish relation types and identify inter-clause relations at all since such information does not exist in Blackboard Treebank.

In addition to the lack of training data, there is also a lack of research into Thai dependency parsing in general. Most notable is the fact that, at the time of writing, there has yet to be any report of systematic application and evaluation of transformer-based dependency parsers for Thai. This gap provides a clear avenue for research into the topic.

In this paper, we aim to partially address this lack of training data and application of state-of-the-art models by 1) introducing a new treebank, called TUD, which consists of 3,627 trees annotated in compliance with the Universal Dependencies (UD) framework \citep{demarneffe2021universal_dependencies}, and 2) training and evaluating newly implemented transformer-based parsers on said treebank. While the architectures we implement might not be novel, our experiments provide the first systematic evaluation of Thai transformer-based dependency parsers and the models serve as good baselines for further research into Thai dependency parsing.

TUD is publicly released on GitHub. The source code used for, and the data generated from, the experiments described in Section \ref{sec:experiment} are also available on the same GitHub repository.

\section{Related work}

\subsection{Universal Dependencies (UD)}

Differences in the set of possible tags and the way to analyze particular constructions of natural languages can lead to differences in parses both within and between languages, thus hindering the development of large-scale and cross-lingual parsers. UD aims to solve this problem by providing a consistent framework for annotating parts-of-speech (POS), morphological features, and syntactic relations across all human languages. The framework was described in their paper \citep{demarneffe2021universal_dependencies}, with examples from many languages. It also encourages the creation of language-specific documentations with examples to clarify how the framework can be applied to each language. For Thai, a manual has already been written by \citet{wirote2023thai_manual}. The manual provides not only clarification of UD principles when applied to Thai but also a summary in Thai of core UD annotation guidelines.

\subsection{Thai dependency parsing}

As stated in the introduction, not much research has been done on Thai dependency parsing and Thai parsers' performance is still far from the state-of-the-art performance achieved in high-resource languages, such as English, for which a result as high as 96.26 LAS has been reported \citep{mrini2020good_english_parser}. Among the few available research papers on Thai dependency parsing, \citet{singkul2019thai_parsing} describes in their paper an experiment in which 7 algorithms are implemented for Thai dependency parsing, 5 transition-based algorithms and 2 graph-based algorithms. However, none of them takes advantage of pretrained transformers and instead uses convolutionl neural networks (CNN) or long short-term memory (LSTM) as encoders, both of which are considered sub-optimal for NLP tasks after the introduction of pretrained transformers \citep{devlin2019bert}. For Thai, 2 large pretrained transformers are publicly available, WangchanBERTa \citep{lowphansirikul2021wangchanberta}, and PhayaThaiBERT \citep{sriwirote2023phayathaibert}. To the best of the authors' knowledge, there is currently only 1 study \citep{yasuoka2023sequence_parsing} that uses a pretrained transformer to perform Thai dependency parsing. However, instead of using WangchanBERTa,\footnote{PhayaThaiBERT cannot be used since it is released after said study has been published.} which has been pretrained on large amount of text, a new transformer pretrained from scratch on Thai Wikipedia, a much smaller data, is used. This means that state-of-the-art Thai dependency parsers, ones that use either WangchanBERTa or PhayaThaiBERT together with either transition- or graph-based algorithms, have not been implemented and evaluated in a published academic paper before.

Another problem with Thai dependency parsing is the fact that many parsers implemented in various studies are strictly research-oriented, with no easy access and usage available for end users. For this reason, many users who need to perform Thai dependency parsing will resort to using more convenient, language-agnostic models that can be trained from scratch to parse any language. Among the most widely-used language-agnostic models are MaltParser \citep{nivre2006maltparser} and UDPipe \citep{straka2016udpipe}.

\section{Treebank creation process}

\subsection{Raw text sources and preprocessing}

Raw text was taken from two sources, the Thai National Corpus (TNC) \citep{wirote2009tnc}, and the November 2020 dump of Thai Wikipedia. Due to the large amount of text available, we randomly select only some documents as data to be annotated while making sure the selected documents are from a wide range of genres, including news articles, Wikipedia articles, essays, advertisement, interviews, and stories.

Each document is split into multiple paragraphs using the newline character as separators. Each paragraph is then tokenized automatically using PyThaiNLP's newmm tokenizer \citep{pythainlp}. Any resulting tokenization errors will be manually corrected during annotation. Note that we do not split each paragraph into sentences. That will be done in the postprocessing step.

\subsection{Data annotation}

Annotation of the treebank was done on Datasaur's labeling platform.\footnote{\url{https://datasaur.ai/}} After annotation is finished, the annotated data will be exported as CoNLL-U format as specified by the UD framework. We recruit a total of 10 annotators, all of which either have at least a bachelor's degree education in linguistics or are in the process of getting one. We use the manual written by \citet{wirote2023thai_manual} as annotation guidelines. The tasks of each annotator are listed below. Any information not specified in the list is not annotated.

\begin{enumerate}
    \item Correct any tokenization errors resulting from the preprocessing step.
    \item For each token, label a Universal POS (UPOS) tag.
    \item For each token, link the token with its head token.
    \item For each dependency link, label the link's Universal Dependency relation type.
\end{enumerate}

The annotators are first instructed to study the annotation manual \citep{wirote2023thai_manual}, after which an instructor will give a lecture explaining the theories and provide parsed examples of various structures expected to be encountered in real data. After that, the annotators will be tasked with annotating a set of pilot sentences. These sentences are specifically designed to cover a wide range of structures explained in the lecture. Once pilot sentences are completely annotated, the instructor will identify common errors in the pilot annotation and re-emphasize what the correct annotation should be. In case of ambiguity, a consensus will be made for each ambiguous case to ensure that the treebank remains consistent. Finally, the annotators go on to annotate real sentences prepared in the previous step. During annotation, the instructor always remains available for questions and occasionally looks at the annotated data to ensure annotation quality. Any perceived errors are then dealt with on a case-by-case basis.

\subsection{Postprocessing and quality control}

After all paragraphs had been annotated, we automatically split them into separate trees based on the annotated dependency links. That is, all connected tokens were extracted as individual trees. Any resulting trees that either 1) contained only a single token, 2) were not completely labeled, or 3) were not valid dependency trees\footnote{For example, trees that contained loops or more than a single root were considered invalid.}, were discarded. After postprocessing, the final treebank consists of 3,627 valid and completely labeled trees.

A final layer of quality control was then done by randomly sampling 50 trees from the treebank and identifying common annotation errors. The identified errors were then targeted for manual correction.

\subsection{Annotation consistency}

To assess annotation consistency, we randomly sampled 20 trees, containing 399 tokens, from the treebank and had them annotated by an expert using the same guidelines that the annotators used. The two sets of annotations, one by the annotators and another by the expert, were then compared to calculate agreement. For POS and dependency relation types, the Cohen's Kappa score is 0.92 and 0.84 respectively, indicating strong agreement. For dependency arcs, we cannot use Cohen's Kappa since the arcs can be arbitrarily long, meaning there is no finite set of labels. Instead, we will calculate 2 simple agreement scores. The first one is the rate at which the arcs are in agreement, regardless of the dependency relation types. The second one is similar to the first one except the dependency relation types must also be in agreement to count toward the agreement score. We will call these scores \textit{unlabeled agreement} and \textit{labeled agreement} respectively. Our unlabeled agreement score is 0.85 and our labeled agreement score is 0.78.

\subsection{Train-dev-test split}

We randomly split TUD into 3 splits, train, dev, and test, using a 8:1:1 ratio. Except for a few less frequent labels, the resulting splits already have similar label distributions. We fix the imbalance in infrequent labels by manually swapping trees between splits until a satisfactory distribution, in which every label is represented equally in each split, is achieved. Various statistics about each split are available on TUD's GitHub repository.

\section{Experiment and evaluation}\label{sec:experiment}

\subsection{Experimental setup}

\subsubsection{Benchmarking models}
\label{sec:models}

Our experiment aims to evaluate the effect different design choices have on the final parsing accuracy. In order to discuss the various components to be evaluated, we will conceptually divide each of our models into 2 main parts: a feature extractor, and a parser.

\textbf{Feature extractors} deal with converting natural language inputs into numeric representations. Each token in the input sentence will be mapped to a set of embeddings that will be concatenated to create the final representation of that token. The embeddings are:
\begin{enumerate}
    \item \textit{Word embedding.} It is computed through the attention mechanism \citep{vaswani2023attention} as part of a transformer's encoding process. The available pretrained encoder-based transformers for Thai are WangchanBERTa \citep{lowphansirikul2021wangchanberta} and PhayaThaiBERT \citep{sriwirote2023phayathaibert}.
    \item \textit{POS embedding.} An embedding is assigned for each possible POS tag.
    \item \textit{Sentence embedding.} Each token's embedding will be concatenated with a copy of the same sentence embedding shared between all tokens of the same sentence, following the method proposed by \citet{altintas2023augmentation}. The embedding is taken from the transformer, which also computes a representation of the entire sentence as part of its encoding process in addition to each token's word embedding.
    \item \textit{Super-token embedding.} Also following the method proposed by \citet{altintas2023augmentation}, each token's embedding will be concatenated with additional ``super-token embeddings'' that represent groups of 2-5 surrounding tokens. These embeddings are obtained by passing convolutional neural network (CNN) filters of varying sizes through the sequence of embeddings obtained prior.
\end{enumerate}

\textbf{Parsers} deal with predicting the sentence's dependency structure given a sequence of token embeddings. They can be traditionally classified into 2 types \citep{jurafsky2024nlp_book}:
\begin{enumerate}
    \item \textit{Graph-based parsers.} We will implement the deep biaffine attention architecture proposed by \citet{dozat2017biaffine}. Our architecture is identical to the one proposed in the original paper with the exception of the Bi-LSTM encoders, where we will use transformer-based encoders instead. A root-constrained algorithm proposed by \citet{zmigrod2020spanning_tree} will be used to extract maximum arborescence from the adjacency matrices produced by the parsers.
    \item \textit{Transition-based parsers.} They can be classified further based on what transition system they use. We will implement 2 transition systems, traditionally called arc-standard, and arc-eager. The oracles in both cases are simple feedforward neural networks that predict an appropriate transition based on the features of 3 tokens. The first 2 tokens are determined by the transition system. The third token is the one that comes immediately after the first 2 tokens.
\end{enumerate}

In order to evaluate the effect each component has on the final performance, we will implement a total of 36 models, each with a different set of components depending on the following design choices. The numbers in parentheses are the numbers of alternatives each choice entails.

\begin{enumerate}
    \item Whether to use an arc-standard transition-based parser, an arc-eager transition-based parser, or a graph-based parser. \textbf{(3)}
    \item Whether to use WangchanBERTa, or PhayaThaiBERT as encoder. \textbf{(2)}
    \item Whether to concatenate sentence embeddings and super-token embeddings into token embeddings as proposed by \citet{altintas2023augmentation}, or not. \textbf{(2)}
    \item Whether to use gold-standard POS tags that come with the treebank, use automatically tagged POS, or not use POS information at all. \textbf{(3)}
\end{enumerate}

As stated above, the number of models to be implemented is 3 x 2 x 2 x 3 = 36. They are trained once for each of the two treebanks used in our experiment (see Section \ref{sec:treebank}). Therefore, a total of 72 models will be trained and evaluated.

In addition to these models, we will also train two existing models to demonstrate that our models can indeed outperform non-transformer-based models, thus making them good baselines for further research. The two models are MaltParser 1.9.2,\footnote{\url{https://www.maltparser.org/}} and UDPipe 1.3.0,\footnote{\url{https://github.com/ufal/udpipe/}} both of which are transition-based parsers that do not use pretrained transformers but are nevertheless widely used to parse Thai sentences due to convenience. In our experiment, these latter 2 models always use gold-standard POS since they are not meant to be serious competitors to other transformer-based models. The default hyperparameters are used for MaltParser since our experiment shows that they are already optimal for Thai data. For UDPipe, the only hyperparameter that differs from the default is the number of training iterations, which is increased from 10 to 20.

\begin{figure*}
    \centering
    \includegraphics[scale=0.7]{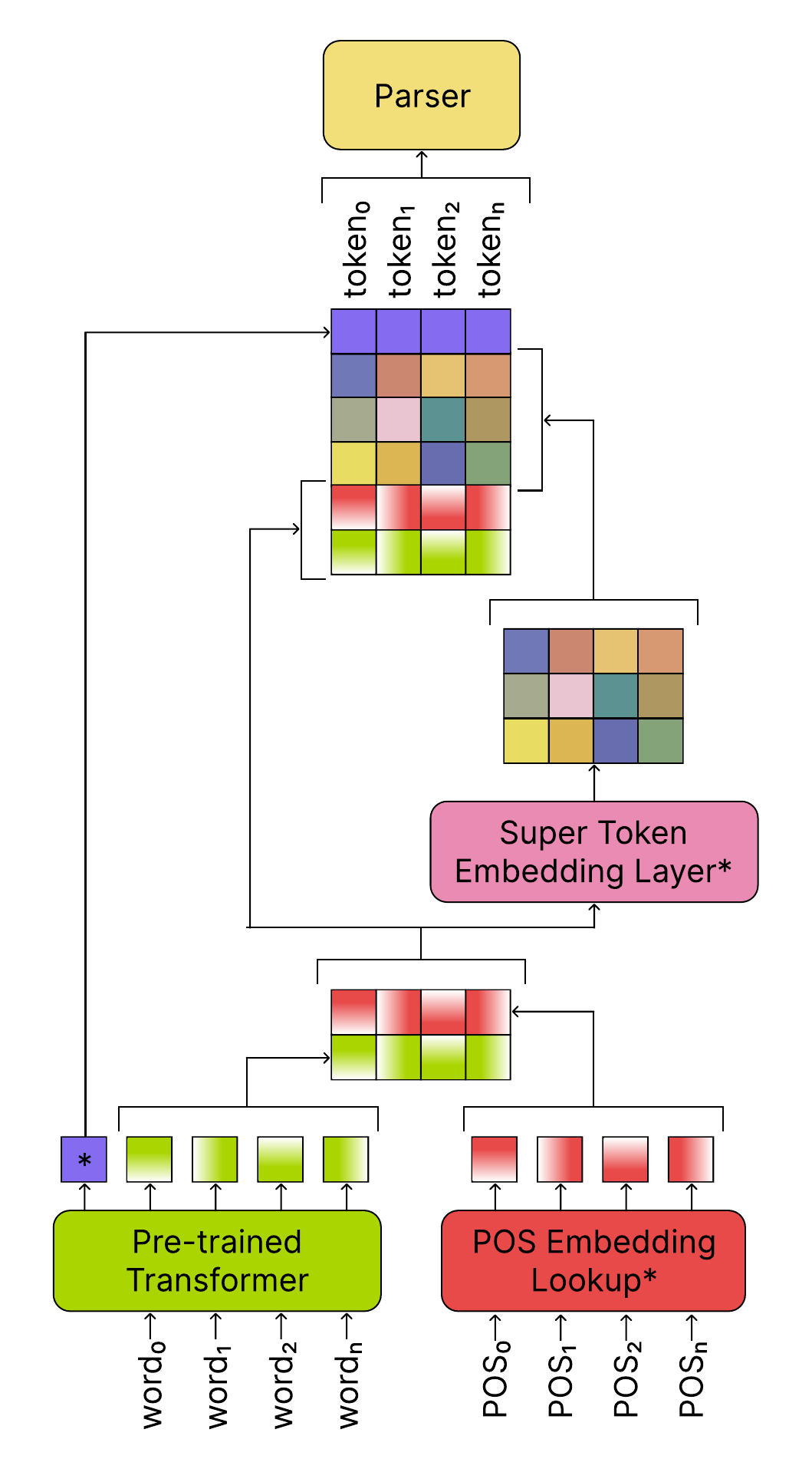}
    \caption{General architecture of the models to be evaluated. Components marked with asterisks (*) are optional.}
    \label{fig:general_architecture}
\end{figure*}

Note that our models use the gold-standard tokenization that comes with the treebank. That is, they never tokenize the input sentences themselves. The scope of this paper deals with parsing and not tokenization. Figure \ref{fig:general_architecture} shows a general diagram of our architecture. Components marked with asterisks (*) are optional. That is, some models might not incorporate them depending on the aforementioned design choices. Table \ref{tab:parser_hyperparameter} shows the hyperparameters used to train said architecture.

\begin{table}
    \begin{tabular}{lc}
        \toprule
        \textbf{Hyperparameter} & \textbf{Value} \\
        \midrule
        No. of parser's hidden layers & 1 \\
        Parser's hidden dim. & 768 \\
        Super-token filter sizes & 2, 3, 4, 5 \\
        Super-token embedding dim. & 192 each \\
        Dropout & 0.1 \\
        Learning rate scheduler & Linear \\
        Warmup ratio & 0.1 \\
        Peak learning rate & 3e-5 \\
        Weight decay & 0.01 \\
        Adam $\epsilon$ & 1e-8 \\
        Adam $\beta_1$ & 0.9 \\
        Adam $\beta_2$ & 0.999 \\
        Batch size & 8 \\
        No. of training epochs & 10 \\
        \bottomrule
    \end{tabular}
    \caption{Hyperparameters used in the training of our models.}
    \label{tab:parser_hyperparameter}
\end{table}

\subsubsection{POS tagger}\label{sec:tagger}

Since POS tagging is outside the scope of this paper, we implement a simple feedforward neural network with no hidden layers that predicts each token's POS based on its embedding obtained from a transformer. The classifier head and the transformer are trained together end-to-end, once for each treebank, using the gold-standard tags that come with the treebank as targets. Once tagged, the POS information of each treebank is shared by every model that does not use gold-standard POS tags.

Our experiment shows that using PhayaThaiBERT as transformer almost always results in better POS accuracy compared to using WangchanBERTa. Therefore, we will only use PhayaThaiBERT for POS tagging. Table \ref{tab:tagger_hyperparameter} shows the hyperparameters used to train our POS tagger.

\begin{table}
    \begin{tabular}{lc}
        \toprule
        \textbf{Hyperparameter} & \textbf{Value} \\
        \midrule
        Transformer & PhayaThaiBERT \\
        Dropout & 0.1 \\
        Learning rate scheduler & Linear \\
        Warmup ratio & 0.1 \\
        Peak learning rate & 3e-5 \\
        Weight decay & 0.01 \\
        Adam $\epsilon$ & 1e-8 \\
        Adam $\beta_1$ & 0.9 \\
        Adam $\beta_2$ & 0.999 \\
        Batch size & 32 \\
        No. of training epochs & 20 \\
        \bottomrule
    \end{tabular}
    \caption{Hyperparameters used in the training of our POS taggers.}
    \label{tab:tagger_hyperparameter}
\end{table}

\subsubsection{Benchmarking treebanks}
\label{sec:treebank}

We will experiment with two publicly available Thai dependency treebanks, \textbf{Thai-PUD} and our \textbf{TUD}. We will not experiment with Blackboard Treebank due to its lack of complete dependency information as stated in the introduction.

Each treebank consists of 3 splits, train, dev, and test. Each split will be used according to the standard machine learning practice and the same splits are used to train both the POS taggers and the dependency parsers. For Thai-PUD, which does not have official train-dev-test splits, we will randomly split it into train, dev, and test splits using the ratio 8:1:1 respectively, which is the same ratio used to split TUD.

\subsubsection{Metrics}

The metrics that will be used to evaluate the models' performance are \textbf{Unlabeled Attachment Score (UAS)} and \textbf{Labeled Attachment Score (LAS)}.

\subsection{Results and discussion}

Table \ref{tab:result} shows the evaluation results of each model on each treebank's test split. Section \ref{sec:previous_models} discusses these results in comparison with previous models. Sections \ref{sec:design_begin} through \ref{sec:design_end} discuss what these results imply about the design choices described in Section \ref{sec:models} using the minimal pair principle. Finally, Section \ref{sec:analysis} discusses availability of each model's full prediction for future analysis.

\begin{sidewaystable*}
    \begin{tabular*}{\textheight}{@{\extracolsep\fill}ccccccccccccc}
        \toprule
        & \multicolumn{6}{c}{Thai-PUD} & \multicolumn{6}{c}{TUD} \\
        \cmidrule{2-7}\cmidrule{8-13}
        Model & \multicolumn{2}{c}{Gold POS} & \multicolumn{2}{c}{Auto POS} & 
        \multicolumn{2}{c}{POS-Agnostic} & \multicolumn{2}{c}{Gold POS} & \multicolumn{2}{c}{Auto POS} & \multicolumn{2}{c}{POS-Agnostic} \\
        \cmidrule{2-3}\cmidrule{4-5}\cmidrule{6-7}\cmidrule{8-9}\cmidrule{10-11}\cmidrule{12-13}
        & UAS & LAS & UAS & LAS & UAS & LAS & UAS & LAS & UAS & LAS & UAS & LAS \\
        \midrule
        MaltParser 1.9.2 & 80.53 & 73.14 & - & - & - & - & 80.56 & 73.66 & - & - & - & - \\
        UDPipe 1.3.0 & 79.7 & 73 & - & - & - & - & 82.34 & 75.28 & - & - & - & - \\
        TSW & 88.14 & 80.39 & 85.28 & 76.65 & 85.6 & 75.45 & 89.47 & 82.6 & 86.27 & 76.22 & 86.59 & 76.81 \\
        TSWA & 88.83 & 82.23 & 88.14 & 80.2 & 86.25 & 76.6 & 89.82 & 83.18 & 86.59 & 76.52 & 86.8 & 76.87 \\
        TEW & 87.4 & 80.53 & 88.0 & 79.6 & 84.54 & 75.03 & 89.2 & 82.27 & 86.33 & 76.53 & 86.02 & 76.02 \\
        TEWA & 88.42 & 81.91 & 87.77 & 80.39 & 86.39 & 78.08 & 89.41 & 82.62 & 86.24 & 76.7 & 86.37 & 76.55 \\
        TSP & 89.57 & 82.33 & 87.91 & 79.51 & 84.73 & 75.27 & \textbf{90.15} & 83.57 & 87.05 & 77.6 & \textbf{87.19} & 77.64 \\
        TSPA & 89.43 & 83.48 & 88.28 & 80.94 & 85.65 & 76.7 & 90.04 & \textbf{83.74} & \textbf{87.26} & 77.55 & 87.09 & \textbf{77.68} \\
        TEP & 89.11 & 82.6 & \textbf{88.92} & 80.48 & 86.48 & 78.17 & 89.93 & 83.42 & 86.82 & 77.09 & 86.54 & 77.07 \\
        TEPA & 89.39 & 83.76 & 88.37 & 81.17 & 87.45 & 79.51 & 89.77 & 83.42 & 87.0 & \textbf{77.68} & 86.76 & 77.61 \\
        GW & 85.97 & 80.43 & 83.43 & 76.6 & 84.36 & 77.34 & 86.33 & 79.64 & 84.25 & 74.59 & 84.77 & 74.41 \\
        GWA & 87.82 & 82.69 & 86.29 & 79.79 & 83.8 & 76.14 & 87.99 & 81.01 & 81.44 & 71.5 & 85.62 & 75.53 \\
        GP & 89.29 & 84.82 & 88.42 & 82.19 & 87.91 & 81.68 & 88.75 & 82.25 & 85.73 & 76.12 & 86.4 & 76.56 \\
        GPA & \textbf{89.8} & \textbf{84.91} & 88.65 & \textbf{82.6} & \textbf{88.74} & \textbf{82.05} & 89.48 & 82.98 & 86.03 & 76.4 & 85.84 & 76.14 \\
        \bottomrule
    \end{tabular*}
    \caption{Evaluation results of each model on each treebank's test split.\\T = Transition-based, S = Arc-standard, E = Arc-eager, G = Graph-based, W = WangchanBERTa, P = PhayaThaiBERT, A = Feature-augmented}
    \label{tab:result}
\end{sidewaystable*}

\subsubsection{Comparison with previous models}
\label{sec:previous_models}

The vast majority of our models achieve higher performance than both MaltParser and UDPipe models. For Thai-PUD, all 36 models, including those that do not use POS information at all, outperform both models in both UAS and LAS. For TUD, only 3 of our 36 models fail to outperform both models in both UAS or LAS.

When compared with models trained and evaluated on Thai-PUD in previous papers, the vast majority of our models also achieve higher performance. The best model reported by \citet{singkul2019thai_parsing} achieves 78.48 UAS (LAS not reported) while our worst model achieves 83.43 UAS, meaning all of our models outperform all models reported in the paper. This is not surprising since none of the models implemented in said paper uses pretrained transformers. Meanwhile, the best transformer-based model reported by \citet{yasuoka2023sequence_parsing} achieves 77.53 LAS (UAS not reported), outperforming 9 of our 36 models, most of which are those that do not use POS information. However, these comparisons might not be perfectly accurate since Thai-PUD does not have an official train-dev-test splits, meaning the splits used in each paper are randomly made and thus not identical.

\subsubsection{WangchanBERTa vs PhayaThaiBERT}
\label{sec:design_begin}

From our 72 models, 36 minimal pairs can be formed to compare the performance of WangchanBERTa and PhayaThaiBERT in different contexts. The results show that models using PhayaThaiBERT as encoders outperform those using WangchanBERTa in UAS in 34/36 pairs and in LAS in 35/36 pairs, strongly suggesting that PhayaThaiBERT is the superior encoder choice for Thai dependency parsing. This supports the results reported in PhayaThaiBERT's paper \cite{sriwirote2023phayathaibert}, in which it is suggested that PhayaThaiBERT has better potential than WangchanBERTa when used in downstream tasks.

\subsubsection{Graph-based vs Transition-based}

We cannot directly form minimal pairs to compare between these 2 architectures since, under our formulation, the transition-based architecture has an additional design choice of what transition system to use. We will go around this problem by selecting, for each standard-eager transition-based model pair, the models with higher LAS as the models that will form minimal pairs with their graph-based counterparts. This process eliminates transition system choice from the context of the minimal pairs, resulting in 24 minimal pairs.

The results show that transition-based models outperform graph-based models in UAS in 19/24 pairs and in LAS in 16/24 pairs. This, however, does not mean that transition-based models are superior. Between models that use the same POS quality, the best model is almost always graph-based for Thai-PUD while it is always transition-based for TUD. These results might suggest that different architectures are appropriate for different data and experiments should still be done to determine which architecture is best for each use case.

\subsubsection{Arc-standard vs Arc-eager}

From our 48 transition-based models, 24 minimal pairs can be formed. Out of these 24 pairs, arc-standard models outperform arc-eager models in UAS in 17/24 pairs. However, contradictorily, arc-standard models outperform arc-eager models in LAS in only 11/24 pairs. These ambiguous results mean that one cannot be considered clearly superior to the other. Nevertheless, they might suggest that the arc-standard transition system has a slight advantage when correct dependency relation types are not desired.

\subsubsection{Efficacy of the feature augmentation method}

From our 72 models, 36 minimal pairs can be formed between augmented and non-augmented models. Out of these 36 pairs, models augmented according to the method proposed by \citet{altintas2023augmentation} outperform non-augmented models in UAS in 26/36 pairs and in LAS in 31/36 pairs, confirming the method's efficacy for Thai data. While the original paper only experiments with graph-based models, our results show that the same method is also beneficial for transition-based models.

\subsubsection{Effect of different POS tag quality}
\label{sec:design_end}

From our 72 models, 24 minimal triplets can be formed. Each triplet consists of 3 models that have the exact same architecture but differ in the quality of POS tags used as features, which can either be gold-standard, automatically tagged, or agnostic. We will call them gold-models, auto-models, and agnostic-models respectively. The last one means the models do not use POS information at all.

The results show that using gold-standard POS almost always leads to superior performance. All gold-models outperform agnostic-models in their triplets and only 1 gold-model fails to outperform an auto-model in its triplet in UAS only i.e. it still achieves the best LAS in its triplet. Meanwhile, auto-models outperform agnostic-models in their triplets in UAS in 14/24 cases and in LAS in 17/24 cases. This might suggest that POS information is more useful in determining dependency relation types and is less useful in predicting untyped dependency arcs.

\subsubsection{Data for further analysis}
\label{sec:analysis}

Due to the large amount and the diversity of data points generated by the 72 models, we will not perform a full error analysis in this paper lest we risk making it excessively long. Instead, we will make the full test split prediction of every model available on TUD's GitHub repository for later investigations.

\section{Conclusion and future work}

In this paper, we create and release a new Thai dependency treebank, named TUD, consisting of 3,627 trees annotated according to the Universal Dependencies (UD) framework. We then implement 36 models, each with slightly different design choices, and train them on Thai-PUD and the newly created TUD. The evaluation results not only show that our models can outperform many previously reported models but also provide insight into the kind of components that should be included in Thai dependency parsers. In particular, they show that 1) PhayaThaiBERT is superior to WangchanBERTa as an encoder in Thai dependency parsing task, 2) there is no clear advantage between using graph-based parsers and transition-based parsers, 3) there is no clear advantage between using arc-standard and arc-eager transition systems in transition-based parsers, 4) the success of the feature augmentation method proposed by \citet{altintas2023augmentation} extends to Thai data and the method also benefits transition-based models in addition to graph-based models, and 5) accurate POS information can significantly improve dependency parsers. The full prediction on test splits of every model we implement is then made publicly available for future analysis.

In future work, we wish to update the Thai UD annotation manual \citep{wirote2023thai_manual} to reflect some of the dependency structures unique to the Thai language that we discovered while annotating the TUD. The TUD will be reannotated and updated at a later date if necessary, should new findings come to light after consultation with linguists.  We also hope to perform extensive error analyses of the models' predictions in order to gain insight into the remaining gaps in Thai dependency parser performance.

\iftaclpubformat

\section*{Acknowledgements}

This work is supported by the National Research Foundation, Singapore under its AI Singapore Programme.

\else

\fi

\bibliography{tacl2021}
\bibliographystyle{acl_natbib}







  
\end{document}